\pdfoutput=1

\documentclass[11pt]{article}

\usepackage[preprint]{coling}

\usepackage{times}
\usepackage{latexsym}
\usepackage{tabularx}
\usepackage{microtype}
\usepackage{enumitem}
\usepackage{xcolor}

\usepackage[T1]{fontenc}

\usepackage[utf8]{inputenc}
\usepackage{amsmath}
\usepackage{graphicx}
\usepackage{amssymb}
\usepackage{dsfont}
\usepackage{algorithm}
\usepackage{algpseudocode}
\usepackage{booktabs}
\usepackage{colortbl}
\usepackage{multirow}
\usepackage{multicol}
\definecolor{color1}{HTML}{F7E8E9}
\definecolor{color2}{HTML}{EF7D8C}
\definecolor{blue1}{HTML}{8AB1EB}
\definecolor{blue2}{HTML}{0CD3FA}

\title{Towards Adaptive Mechanism Activation in Language Agent}

\author{Ziyang Huang, Jun Zhao, Kang Liu \\
        The Key Laboratory of Cognition and Decision Intelligence for Complex Systems, \\
        Institute of Automation, Chinese Academy of Sciences \\
        School of Artificial Intelligence, University of Chinese Academy of Sciences \\
    \texttt{huangziyang2023@ia.ac.cn}
}

\begin{document}
\maketitle
\begin{abstract}
Language Agent could be endowed with different mechanisms for autonomous task accomplishment. Current agents typically rely on fixed mechanisms or a set of mechanisms activated in a predefined order, limiting their adaptation to varied potential task solution structures. To this end, this paper proposes \textbf{A}daptive \textbf{L}anguage \textbf{A}gent \textbf{M}echanism \textbf{A}ctivation Learning with Self-Exploration (\textbf{ALAMA}), which focuses on optimizing mechanism activation adaptability without reliance on expert models. Initially, it builds a harmonized agent framework (\textbf{UniAct}) to \textbf{Uni}fy different mechanisms via \textbf{Act}ions. Then it leverages a training-efficient optimization method based on self-exploration to enable the UniAct to adaptively activate the appropriate mechanisms according to the potential characteristics of the task. Experimental results demonstrate significant improvements in downstream agent tasks, affirming the effectiveness of our approach in facilitating more dynamic and context-sensitive mechanism activation.
\end{abstract}

\section{Introduction}
Language Agent (LA) \cite{sumers2024cognitive, yao2023react, xi2023rise, gao2023large} has garnered considerable attention recently due to the rapid advancements in Large Language Models (LLMs) \cite{gpt4o, llama3modelcard, yang2023baichuan, chowdhery2022palm, radford2018improving}. Through the well-designed prompts and carefully selected in-context demonstrations \cite{zhou2024selfdiscover, dong2023survey, liu2021pretrain},  LLM-based agents can be endowed with different mechanisms\footnote{Here, \textbf{mechanism} is defined as the inherent ability of the Language Agent which could be manifested as a special workflow externally and activated by the specific prompting.} for environment interaction and task solving. Existing LAs could benefit from distinct mechanisms for various tasks with unique solution structures \cite{zhou2024selfdiscover}. For instance, Reflexion \cite{shinn2023reflexion} is equipped with \texttt{Reflection} mechanism to gain insightful refinement suggestions. And ReAct \cite{yao2023react} is equipped with \texttt{External-Augmentation} mechanism to ground the solution trajectory with additional evidence.

\begin{figure}[t]
    \centering
    \includegraphics[width=\linewidth]{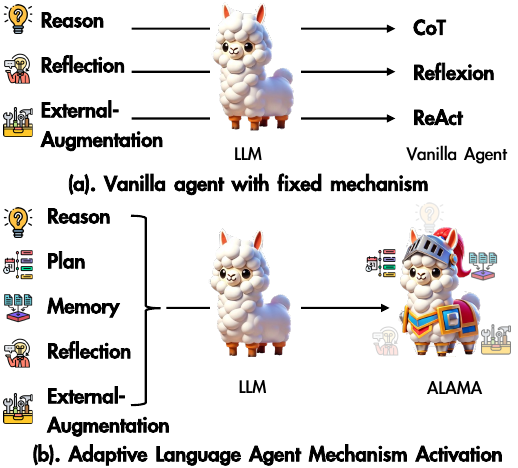}
    \caption{Illustration of Language Agent with different mechanisms. (a). Vanilla agent with fixed mechanisms by In-Context learning. (b). ALAMA with different mechanisms learn to fit into different environments by Self-Exploration.}
    \label{fig:main}
\end{figure}

Despite the success of current LAs through aforementioned direct prompting and in-context learning, named as \textit{manual mechanism activation}, they rely on fixed mechanisms or a predefined sequence of mechanisms \cite{liu-etal-2023-plan, chen2023fireact, song2024trial}, as illustrated in Figure \ref{fig:main} (a). As a result, such rigidity hampers activating the optimal solution structures (mechanism) for a specific task and also limits their adaptability to open-world scenarios. There is compelling evidence that \textit{oracle language agent mechanism activation}, selecting the most appropriate mechanism for a task, can improve the performance by over 15\% compared to fixed mechanism baselines (as shown in Section \ref{sec:4.1}). Therefore, it highlights the significant potential of \textit{adaptive mechanism activation}, the focus of the paper, where mechanisms are adaptively activated based on task characteristics, as shown in Figure \ref{fig:main} (b). We view this as a critical kind of meta-ability for LAs, and its enhancement could offer the potential for better generalization in unseen tasks.

Intuitively, when humans encounter unfamiliar tasks, they tend to first explore different solution strategies and then select the most effective solution from previous experiences in similar tasks. Inspired by this, to enable LAs to adaptively select suitable solution strategies (adaptive mechanism activation), this paper proposes \textbf{A}daptive \textbf{L}anguage \textbf{A}gent \textbf{M}echanism \textbf{A}ctivation Learning with Self-Exploration (\textbf{ALAMA}), a novel technique for learning adaptive mechanism activation across various tasks. It first introduces a harmonized agent framework to \textbf{Uni}fy existing known mechanisms by \textbf{Act}ions (\textbf{UniAct}). Compared with previous agents which did not fully integrate various mechanisms \cite{yao2023react} or only implicitly incorporated specific mechanisms into the thinking process without an explicit trigger \cite{zhou2023leasttomost}, UniAct defines the workflows of mechanisms as specific actions. In this way, different mechanisms would share a unified action space. When the agent triggers an action, the corresponding mechanism is expected to be activated. 

Secondly, to fulfill adaptive mechanism activation in LAs, our ALAMA adopts a self-exploration fine-tuning way rather than simply prompting. Sufficient high-quality trajectories with different activated mechanisms are important for model training but not easy to acquire. To this end, ALAMA firstly leverages self-exploration to obtain sufficient trajectories for training. Compared with previous methods of acquiring trajectories through manual annotation or distillation from proprietary models \cite{zeng2023agenttuning, chen2023fireact}, self-exploration could extremely decrease data acquisition costs and alleviate the paucity of training signals. Specifically, we manually activate different mechanisms to facilitate multiple rounds of self-exploration. Consequently, diverse solution trajectories are produced and then converted into the UniAct format. To introduce implicit mechanism preferences towards different tasks as well as fundamental interaction and instruction-following capabilities for the agent, this paper utilizes \textbf{I}mplicit \textbf{M}echanism \textbf{A}ctivation \textbf{O}ptimization (\textbf{IMAO}), which samples subset of positive trajectories to fine-tune the LAs.

For further model training, different from existing exploration-based methods which use success-failure pairwise data for behavior contrastive learning \cite{song2024trial, yuan2024advancing}, this paper employs \textbf{M}echanism \textbf{A}ctivation \textbf{A}daptability \textbf{O}ptimization (\textbf{MAAO}) based on KTO algorithm \cite{ethayarajh2024kto}. KTO is a preference learning algorithm that only requires binary signals (desirable/ undesirable). In this way, the need for assembling high-quality pairwise data \cite{rafailov2023direct, xie2024montecarlotreesearch} is alleviated and all trajectories with different rewards obtained during the self-exploration phase could be utilized, which makes training more efficient.

To validate the effectiveness of our proposed method, the paper conducts extensive experiments on mathematical reasoning \cite{cobbe2021training, mishra2022numglue, patel-etal-2021-nlp} and knowledge-intensive reasoning \cite{yang-etal-2018-hotpotqa, joshi-etal-2017-triviaqa, press-etal-2023-measuring} tasks. The results show that ALAMA surpasses the baselines by a large margin on both Held-in and Held-out datasets, demonstrating its strong performance and generalization.

To sum up, this paper presents the adaptive mechanism activation as a crucial meta-ability for Language Agents. Moreover, this paper proposes \textbf{A}daptive \textbf{L}anguage \textbf{A}gent \textbf{M}echanism \textbf{A}ctivation Learning with Self-Exploration (\textbf{ALAMA}), which incorporates a harmonized agent framework to unify different mechanisms and a training-efficient optimization method based on self-exploration.


   

\begin{figure*}[htb]
    \centering
    \includegraphics[width=\linewidth]{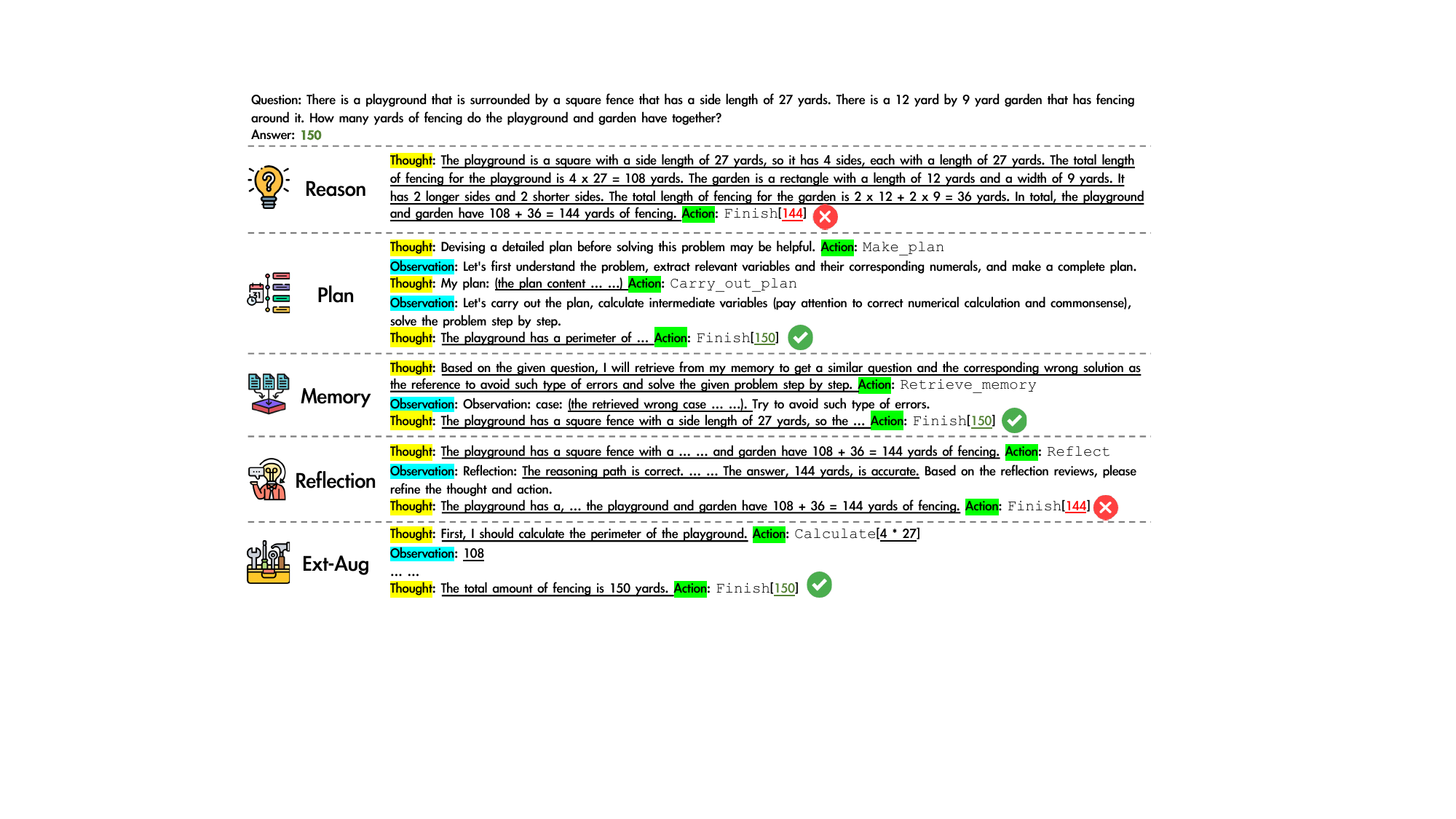}
    \caption{The UniAct trajectory examples for five mechanisms. The underlined contents are generated by the vanilla agent or from external feedback.}
    \label{fig:example}
\end{figure*}

\section{Background}
With different prompts and demonstrations, the agent can be equipped with different mechanisms for better task-solving performance.  This paper selects five essential agent mechanisms as the focus of our study: (1) \texttt{Reason} \cite{wei2022chain}: Directly obtaining the answer through step-by-step reasoning. (2) \texttt{Plan} \cite{wang-etal-2023-plan, zhou2023leasttomost}: First understanding the task and develop a plan to decompose it into smaller, more easily solvable sub-tasks, and then progressively solving each sub-task to arrive at the final answer. (3) \texttt{Memory} \cite{sun2023expnote, gao2024retrievalaugmented}: Initially building a database of failed examples. During each subsequent task execution, similar cases are retrieved from this database based on task similarity (cosine of task description embedding), and the agent could try to avoid similar errors. (4) \texttt{Reflection} \cite{shinn2023reflexion, madaan2023selfrefine}: Introducing a Critic Model into the environment to reflect on the previously reasoned answers by the agent when necessary. (5) \texttt{External-Augmentation} \cite{yao2023react, schick2023toolformer}: Calling task-specific toolkits for solving different tasks, such as a calculator for mathematical reasoning or a search engine for knowledge-intensive reasoning. As shown in Figure \ref{fig:example}, we demonstrate the examples of trajectories with different mechanisms in UniAct format\footnote{We describe the UniAct format and how to transform the agent trajectories into it in Section \ref{sec:alama}.}. We defer the implementation details of each mechanism to the appendix \ref{implementation}.


\begin{figure*}[htb]
    \centering
    \includegraphics[width=\linewidth]{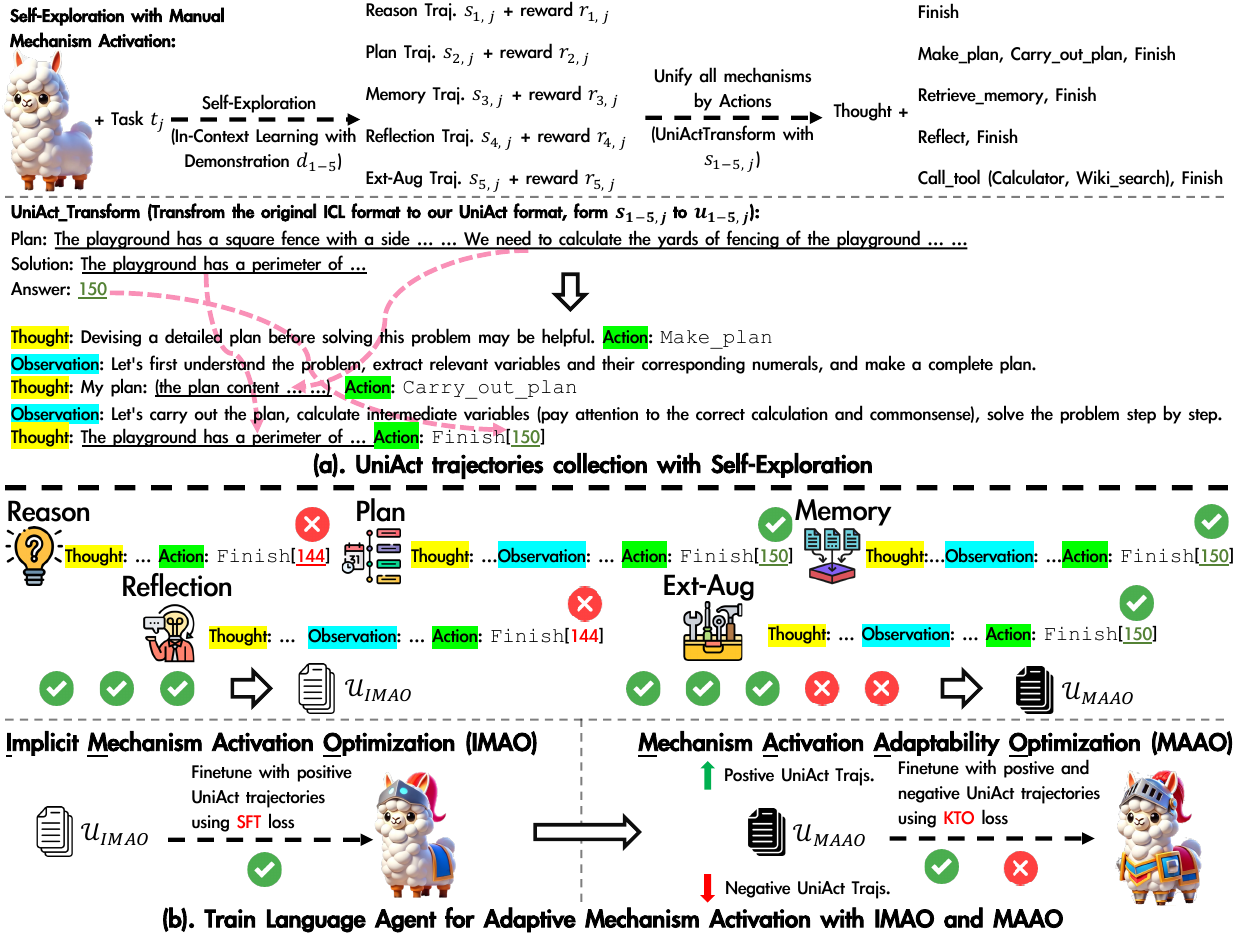}
    \caption{The illustration of ALAMA process. The UniAct trajectories are collected by Self-Exploration with manual mechanism activation. For tasks with mechanism sensitivity, we use the corresponding positive trajectories for Implicit Mechanism Activation Optimization, and utilize both positive and negative ones for Mechanism Activation Adaptability Optimization. }
    \label{fig:method}
\end{figure*}

\section{ALAMA: Adaptive Language Agent Mechanism Activation Learning with Self-Exploration}
\label{sec:alama}
This section describes our method in detail. Firstly, we introduce a harmonized agent framework to unify existing known mechanisms (\textbf{UniAct}). Secondly, we elaborate on a self-exploration fine-tuning method for enhancing the meta-ability of adaptive mechanism activation. In specific, we leverage \textbf{Self-Exploration} with manual mechanism activation to sample various UniAct trajectories. Next, we employ Implicit Mechanism Activation Optimization (\textbf{IMAO}) and Mechanism Activation Adaptability Optimization (\textbf{MAAO}) to adapt the agent to different tasks based on the recognized characteristics and potential solution structures.

\paragraph{UniAct: Unify Agent Mechanisms by~Actions}
Currently, ReAct \cite{yao2023react} serves as the foundational framework for LLM-based agents, employing the \texttt{Thought}, \texttt{Action}, \texttt{Observation} ($\tau, a, o$ as the abbreviation) format to govern agent control. This format only unifies reasoning, action generation, and the acquisition of feedback from external environments into natural language space.  Based on this, we propose UniAct to integrate diverse mechanisms into a unified framework explicitly. As depicted in the upper of Figure \ref{fig:method} (a), we define distinct \texttt{Action}s for each mechanism to unify the different workflows into a shared action space. Specifically, we define $\mathtt{make\_plan}$ for detailed plan generation, $\mathtt{Carry\_out\_plan}$ for plan execution, $\mathtt{Retrieve\_memory}$ to get potential error cases, $\mathtt{Reflect}$ to get insightful correction suggestions from the expert Critic model, $\mathtt{Call\_tool}$ to invoke external tools, and $\mathtt{Finish}$ to output the final results and terminate the solution trajectories. We take the \texttt{Thought} as the thinking process before generating the actions, and the \texttt{Observation} as the environmental feedback. Furthermore, we have adapted the external environment to not only provide task-related feedback but also return appropriate prompt to facilitate the activation of corresponding mechanisms. Details regarding the UniAct format including the actions and corresponding grounding prompts are provided in Appendix \ref{sec:prompt}.


\paragraph{Self-Exploration}
We refer to the base agent with parameter $\theta$ as $\textrm{LA}_{\theta}$ and all the mechanisms as $ \mathcal{M} = \{m_i\}_{i=1}^{N} $. We construct a demonstration trajectory $d_i$ where only that specific mechanism $m_i$ is activated. As shown in upper of Figure \ref{fig:method} (a), given Tasks $\mathcal{T} = \{t_j\}_{j=1}^{|\mathcal{T}|}$, we manually activate different mechanisms by prompting with the corresponding $d_i$ to get the exploration solution trajectory $s_{i,j}$ and corresponding reward $r_{i,j}$. And then we transform all these trajectories into UniAct format $u_{i,j}$. 
\begin{align}
    s_{i,j}, r_{i,j} &= \textrm{LA}_{\theta}(d_i, t_j) \\ 
    u_{i,j} &= \textrm{UniActTransform}(s_{i,j}) \nonumber \\
    &= (\tau_1, a_1, o_1, \cdots, o_{m-1}, \tau_m, a_m)_{i,j} 
\end{align}
where $\tau, a, o$ represent thought, action, and observation respectively. For UniActTransform($\cdot$), we extract the self-generated contents and external feedback from the ICL solutions, and then fill them into the UniAct format with explicit actions. As depicted in the bottom part of Figure \ref{fig:method} (a), we show a transformation example of \texttt{Plan}. Please refer to the Appendix \ref{app:transform} for other mechanisms. Finally, we get all self-exploration UniAct trajectories $\mathcal{U}$.

{
\fontsize{10}{0} \selectfont
\begin{align}
    \mathcal{U} = \{U_j\}_{j=1}^{|\mathcal{T}|} = \{  \{u_{i,1}\}_{i=1}^{N}, \cdots, \{u_{i,|\mathcal{T}|}\}_{i=1}^{N} \}
\end{align}
}

Notably, not every mechanism could fit all tasks and obtain correct results. As illustrated in the upper of Figure \ref{fig:method} (b), certain tasks are successfully solved by specific mechanisms, while remaining unsolved when the other are activated. We refer to these as \textbf{tasks with mechanism sensitivity}.
\paragraph{IMAO: Implicit Mechanism Activation Optimization}
To equip the model with the basic ability to follow the UniAct format in the zero-shot setting and adaptively activate appropriate mechanisms, we sample a subset of positive trajectories from $\mathcal{U}$ for supervised fine-tuning, as shown in the left bottom part of Figure \ref{fig:method} (b). To introduce implicit preferences for different mechanisms across tasks, we use the UniAct trajectories with $r = 1$ of the tasks with mechanism sensitivity as the training data, referred to as $\mathcal{U}_{\textrm{IMAO}}$.

Thoughts and actions are generated by the vanilla agent, while observations are gathered from the environment. Consequently, we compute the next token prediction loss on thought $\tau$ and action $a$, while masking the loss on observation $o$.

{
\fontsize{8}{0} \selectfont
\begin{align}
    &\mathcal{L}_{\textrm{IMAO}}(\textrm{LA}_{\theta}) = \mathbb{E}_{u \in \mathcal{U}_{\textrm{IMAO}}}-\log P(u|t) \\
    &= \mathbb{E}_{u \in \mathcal{U}_{\textrm{IMAO}}}- \log P(a_m, \tau_m, \cdots, a_1, \tau_1 | t) \\
    &= \mathbb{E}_{u \in \mathcal{U}_{\textrm{IMAO}}} \bigg[ - \sum_{k=1}^{m} \log P(\tau_k| o_{k-1}, a_{k-1}, \cdots, t)  \nonumber\\
   &\quad\quad\quad\quad - \sum_{k=1}^{m} \log P(a_k| \tau_k, o_{k-1}, \cdots, t) \bigg]
\end{align}
}
where the $t$ and $u$ represent the task and the corresponding self-generated trajectory.

\paragraph{MAAO: Mechanism Activation Adaptability Optimization}
For all tasks with mechanism sensitivity, we collect all corresponding trajectories as training data, referred to as $\mathcal{U}_{\textrm{MAAO}}$. We treat those with a reward equal to 1 as $\mathcal{U}_{\textrm{MAAO-pos}}$, and the other as $\mathcal{U}_{\textrm{MAAO-neg}}$. Instead of only using positive trajectories in IMAO, our MAAO utilizes the contrastive information between positive and negative examples to update the agent using KTO loss \cite{ethayarajh2024kto}. The behavior of the agent is biased towards positive examples and away from negative ones. This approach enhances the model's meta-ability for adaptive mechanism activation:

{
\fontsize{8}{0} \selectfont
\begin{align}
 &z_0 = \mathbb{E}_{t^\prime \in \mathcal{U}_{\textrm{MAAO}} } [\textrm{KL}(\textrm{LA}_{\theta}(u^\prime|t^\prime) || \textrm{LA}_{\textrm{ref}}(u^\prime|t^\prime))] \\
    & v(t, u) = (-1)^{\mathds{1}(u \in \mathcal{U}_{\textrm{MAAO-pos}})}  \lambda_{\textrm{pos/neg}} \times \nonumber \\
    & \quad\quad\quad\quad\quad\quad \sigma\left(\beta\left(z_0 - \log \frac{\textrm{LA}_{\theta}(u|t)}{\textrm{LA}_{\textrm{ref}}(u|t)}\right)\right) \\
    & \mathcal{L}_{\textrm{MAAO}}(\textrm{LA}_{\theta}, \textrm{LA}_{\textrm{ref}}) =  \mathbb{E}_{u \in \mathcal{U}_{\textrm{MAAO}}} [\lambda_{\textrm{pos/neg}} - v(t, u)] 
\end{align}
}
When $u \in \mathcal{U}_{\textrm{MAAO-pos}}$, $(-1)^{\mathds{1}(u \in \mathcal{U}_{\textrm{MAAO-pos}})} = -1$ , $\lambda_{\textrm{pos/neg}} = \lambda_{\textrm{pos}}$, and vice versa. 

The pseudo-code of the optimization method is shown in Algorithm \ref{algo:alama}.

\section{Experiment}

\begin{table*}[t]
\centering
\tabcolsep=15pt
\resizebox{\linewidth}{!}{
\begin{tabular}{lcccccc}
\toprule
\multirow{3}{*}{} & \multicolumn{3}{c}{\textbf{Mathematical Reasoning} (Acc)} & \multicolumn{3}{c}{\textbf{Knowledge-intensive Reasoning} (EM)}  \\
\cmidrule(l){2-4} \cmidrule(l){5-7}
 & Held-in & \multicolumn{2}{c}{Held-out} & Held-in & \multicolumn{2}{c}{Held-out}  \\
\cmidrule(l){2-2} \cmidrule(l){3-4} \cmidrule(l){5-5} \cmidrule(l){6-7}
 & GSM8K & NumGLUE &SVAMP & HotpotQA & TriviaQA & Bamboogle  \\
\midrule
\multicolumn{2}{l}{GPT-3.5-turbo (one-shot Activation)} \\
\cmidrule(l){1-1}
\texttt{Reason} & 63.91  & 60.63 & 71.20 & 22.20 & 28.80 & 28.80 \\
\texttt{Plan} & 77.94 & 59.84 & 83.40 & 22.80 & 51.20 & 37.60 \\
\texttt{Memory} & 76.42 & 65.75 & 81.10 & 25.80 & 55.60 & \underline{44.80} \\
\texttt{Reflection} & \underline{79.38} & 66.14 & \underline{86.10} & \underline{30.80} & \underline{60.80} & 41.60 \\
\texttt{External-Augmentation} & 70.66 & \underline{70.47} & 79.00 & 22.20 & 44.00 & 30.40\\
\texttt{Average} & \cellcolor{color1}73.66 &\cellcolor{color1}64.57 &\cellcolor{color1}80.16 &\cellcolor{color1}24.76  &\cellcolor{color1}52.16 &\cellcolor{color1}36.64\\
\texttt{Majority Voting} & \cellcolor{blue1}82.25 &\cellcolor{blue1}66.54 & \cellcolor{blue1}86.30 &\cellcolor{blue1}28.40  &\cellcolor{blue1}56.00 &\cellcolor{blue1}41.60\\

\midrule
\multicolumn{2}{l}{Llama-3-8B-Instruct (one-shot Activation)} \\
\cmidrule(l){1-1}
\texttt{Reason} & 73.08 & 41.73 & 66.10 & 17.60 & 41.40 & 29.60 \\
\texttt{Plan} & 77.56 & 68.11 & 82.90 & 19.80 & 44.40 & 31.20 \\
\texttt{Memory} & 77.03 & 70.47 & 77.80 & 16.20 & 41.20 & 30.40\\
\texttt{Reflection} & \underline{80.06} & \underline{74.40} & \underline{85.90} & \underline{26.00} &  \underline{55.80} & \underline{37.60} \\
\texttt{External-Augmentation} & 71.80 & 61.02 & 75.80 & 15.80  & 38.60 & 20.80\\
\texttt{Average} & \cellcolor{color1}75.90 & \cellcolor{color1}63.15 &\cellcolor{color1}77.70 &\cellcolor{color1}19.08  &\cellcolor{color1}44.28 & \cellcolor{color1}29.92 \\
\texttt{Majority Voting} & \cellcolor{blue1}82.71 &\cellcolor{blue1}70.87 &\cellcolor{blue1}85.50 &\cellcolor{blue1}21.60  &\cellcolor{blue1}48.60 &\cellcolor{blue1}37.60 \\

\midrule
\multicolumn{2}{l}{\textbf{ALAMA}} \\
\cmidrule(l){1-1}
\texttt{IMAO} & 78.77 & 72.83 & 83.30 & 24.00  & 40.40 & 27.20\\
\texttt{IMAO + MAAO} & \cellcolor{color1}\textbf{82.18} &\cellcolor{color1}\textbf{78.35} &\cellcolor{color1}\textbf{88.20} & \cellcolor{color1}\textbf{27.60}  &\cellcolor{color1}43.60 & \cellcolor{color1}\textbf{32.80} \\
\texttt{Self-Adapt Consistency} & \cellcolor{blue1}\textbf{85.06} &\cellcolor{blue1}\textbf{79.13} &\cellcolor{blue1}\textbf{89.80} &\cellcolor{blue1}\textbf{31.00}  &\cellcolor{blue1}\textbf{49.40} & \cellcolor{blue1}36.80\\ 
\bottomrule
\end{tabular}
}
\caption{Performance of different methods. We use Accuracy and EM as metric for Mathematical Reasoning and Knowledge-intensive Reasoning.}
\label{tab:main}
\end{table*}

\subsection{Setup}

\paragraph{Model and Datasets} We utilize GPT-3.5-turbo-0125 as the closed-source model baseline, accessed through the OpenAI API. We employ Meta-Llama3-8B-Instruct as the backbone for ALAMA. For datasets, the paper employs the GSM8K \cite{cobbe2021training} and HotpotQA \cite{yang-etal-2018-hotpotqa} as Held-in tasks for exploration, training, and testing. Additionally, we select NumGLUE \cite{mishra2022numglue}, SVAMP \cite{patel-etal-2021-nlp}, TriviaQA \cite{joshi-etal-2017-triviaqa}, and Bamboogle \cite{press-etal-2023-measuring} as Held-out tasks to evaluate the generalization performance. For dataset processing details, please refer to Appendix \ref{app:data}.

\paragraph{Baselines} We select the following baselines for comparisons, like (1) Fixed single mechanism (\texttt{Reason}, \texttt{Plan}, \texttt{Memory}, \texttt{Reflection} and \texttt{External-Augmentation} shown in Table \ref{tab:main}): we manually construct one in-context demonstration example to activate different mechanisms (2) \texttt{Average}: The average performance of different mechanisms. (3) \texttt{Majority Voting}: Selecting the most consistent \cite{wang2023selfconsistency} answer among the solutions obtained by activating different mechanisms as the final answer. (4) \texttt{Self-Adapt Consistency}: We apply self-consistency \cite{wang2023selfconsistency} technique to ALAMA. For training and inference details, please refer to Appendix \ref{app:train}.

\subsection{Main Results}


\paragraph{Adaptive Mechanism Activation outperforms fixed Manual Mechanism Activation.}
As shown in Table \ref{tab:main}, ALAMA outperforms all single mechanism baselines and the average performance of different mechanisms on the Held-in tasks. We consider the Average as the bottom performance for introducing multiple mechanisms into one agent. After IMAO (supervised learning), ALAMA surpasses the Average by 2.87 on GSM8K and 4.92 on HotpotQA, indicating that it has the ability to adaptively activate different mechanisms based on the task characteristics.


Furthermore, after MAAO (preference learning), ALAMA continues to improve by 3.41 on GSM8K and 3.60 on HotpotQA. This suggests that MAAO can enhance the adaptability of the agent to potential solution structures of different tasks. Behavior contrastive learning enables the model to preferentially activate certain specific mechanisms while refusing to activate the remaining ones. For example, in manual activation, \texttt{Plan} outperforms \texttt{Reason} by 4.48 on GSM8K. After MAAO, when the agent encounters specific complex mathematical reasoning tasks that can not be solved directly through reasoning, it recognizes that direct reasoning may lead to incorrect answers and thus chooses to analyze the sub-problems in the question first, decompose the problem, and solve them individually, ultimately summarizing the answers. ALAMA based on Llama-3-8B-Instruct outperforms GPT-3.5-turbo average on Held-in tasks after ALAMA, demonstrating the superior effectiveness.

Compared to all fine-tuning baselines shown in the upper of Table \ref{tab:finetune}, the introduction of multiple mechanisms in ALAMA demonstrates significant performance gains, which adequately exemplifies the superiority of adaptive mechanism activation learning techniques.

\begin{table}[ht]
    \centering
    \resizebox{1\linewidth}{!}{
    \begin{tabular}{lc}
    \toprule
     Agent    & GSM8K (Acc) \\
     \midrule
     Fine-tuning Baselines \\
     \cmidrule{1-1}
     FireAct$_{\textrm{Llama-2-7B}}$ \cite{chen2023fireact}  & 56.10 \\
     Lumos$_{\textrm{Llama-2-7B}}$ \cite{yin2024lumos}   & 54.90 \\
     WizardMath$_\textrm{Llama-2-13B}$ \cite{luo2023wizardmathempoweringmathematicalreasoning} & 63.90 \\	
     ToRA$_\textrm{Llama-2-13B}$ \cite{gou2024toratoolintegratedreasoningagent} & 72.70 \\	
     Husky$_{\textrm{Llama-2-13B}}$ \cite{kim2024husky} & 79.40 \\
     Husky$_{\textrm{Llama-3-8B}}$ \cite{kim2024husky} & 79.90 \\
     MAmmoTH2-8B$_\textrm{Llama-3-8B}$ \cite{yue2024mammoth2scalinginstructionsweb} & 70.40 \\	
     MAmmoTH2-8B-Plus$_\textrm{Llama-3-8B}$ \cite{yue2024mammoth2scalinginstructionsweb} & 84.10 \\	
     \midrule
     Train on Self-Exploration Data \\
     \cmidrule{1-1}
    ALAMA$_{\textrm{Llama-3-8B-SFT}}$ & 78.77 \\
    ALAMA$_{\textrm{Llama-3-8B-DPO}}$ & 80.52 \\
    ALAMA$_{\textrm{Llama-3-8B-KTO}}$ & \textbf{82.18} \\
     \bottomrule
    \end{tabular}
    }
    \caption{Fine-tuning based Language Agent performance comparison. ALAMA with multiple mechanisms optimized with efficient adaptive learning using less data demonstrates suprior performance.}
    \label{tab:finetune}
\end{table}

\paragraph{ALAMA outperforms SoTA fine-tuning baselines with more efficient data acquisition and training.} 

The agent data employed for fine-tuning baselines as presented in Table \ref{tab:finetune} are all curated by expert models or humans.
However, our ALAMA surpasses these baselines merely by relying on self-exploration, which is more efficient. More specifically, \textit{Husky} is trained on agent trajectories from 10 datasets including GSM8K, MATH, and TabMWP. SoTA agent \textit{Mammoth2-Plus} first collects over 10 million instruction data using a complicated pipeline to enhance the reasoning ability and then uses math instruction datasets (including GSM8K and MATH) for supervised fine-tuning. Our ALAMA$_{\textrm{Llama-3-8B-KTO}}$ uses only GSM8K for training. Despite having much more training data, \textit{Husky} underperforms and \textit{Mammoth2-Plus} is only about 2\% higher in performance than ALAMA$_{\textrm{Llama-3-8B-KTO}}$, fully demonstrating the data efficiency of ALAMA.

In addition, we introduced a DPO~\cite{rafailov2023direct} based counterpart, i.e. ALAMA$_{\textrm{Llama-3-8B-DPO}}$. The positive and negative trajectories in $\mathcal{U}_{\textrm{MAAO}}$ are then paired into multiple preference pairs for DPO training. This pairing approach leads to increased training costs. Experiment results demonstrate that KTO yields better results, further highlighting the efficiency and effectiveness of our method.

%


\paragraph{ALAMA demonstrates superior generalization on Held-out tasks.}
Apart from testing on the Held-in datasets, we have also selected four Held-out datasets for evaluation under the zero-shot setting. On NumGLUE and SVAMP, ALAMA outperforms the best baseline by 3.95 and 2.3, respectively. With the assistance of Self-Adapt Consistency, ALAMA surpasses 4.73 and 3.9, respectively. Additionally, ALAMA also outperforms most baselines, including Average, on TriviaQA and Bamboogle. This adequately demonstrates the effectiveness and generalization of our proposed method.

\paragraph{Self-Adapt Consistency outperforms manual mechanism activation based Majority Voting.}
On GSM8K, the performance obtained by selecting the majority answer from the different mechanisms significantly surpasses the performance of all individual mechanisms as well as the average performance. We consider this as a strong baseline for the comprehensive utilization of multiple mechanisms. For a fair comparison, we sample 5 times for Self-Adapt consistency. It exceeds the above strong baseline by 2.35 and 9.4 on GSM8K and HotpotQA respectively, indicating that the fine-tuned ALAMA possesses the ability to adaptively activate different mechanisms. With the help of random sampling, ALAMA activates the most effective task-specific mechanisms to generate diverse trajectories, ultimately achieving better performance.

\section{Analysis}

\subsection{The Specificity of Different Mechanisms}
\label{sec:4.1}
\begin{figure}[tbp]
    \centering
    \includegraphics[width=\linewidth]{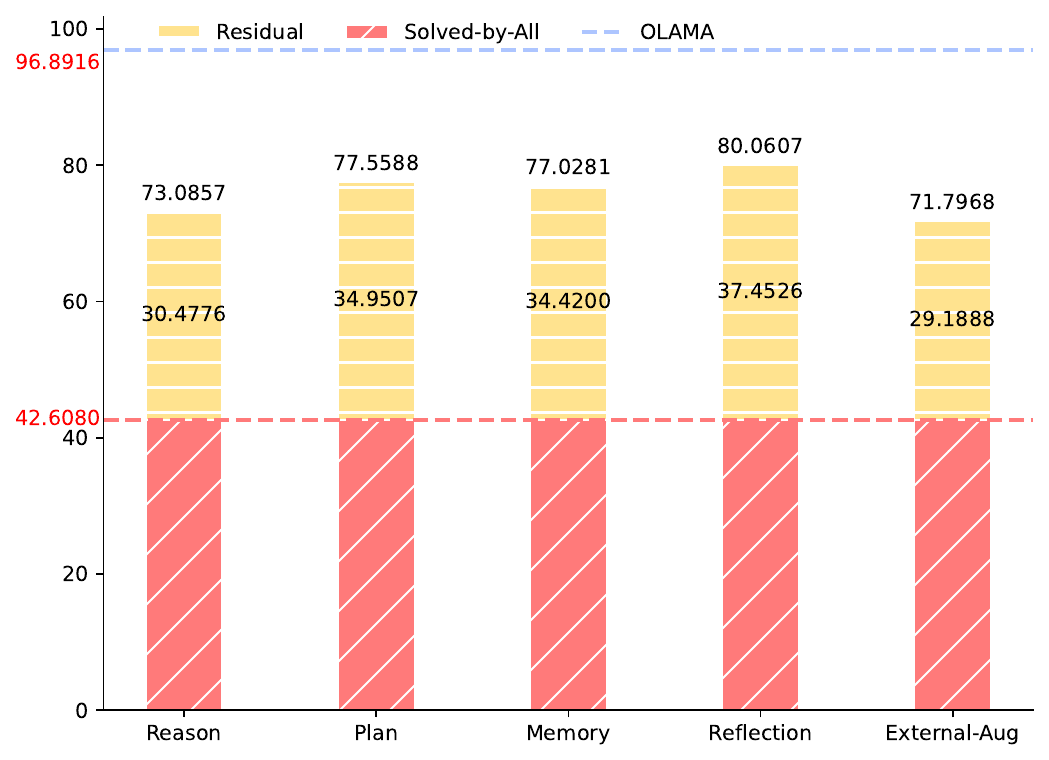}
    \caption{Mechanism specificity analysis results on GSM8K. \texttt{OLAMA} represents oracle mechanism activation, which selects the most appropriate mechanism for each task. \texttt{Solved-by-All} represents that corresponding tasks could be solved by all mechanisms respectively. And \texttt{Residual} represents the performance gap (yellow part) between different mechanisms and \texttt{Solved-by-All}, which shows the specificity.}
    \label{fig:specificity}
\end{figure}

This subsection tries to investigate the impact of different mechanisms on downstream task performance. In detail, we choose Llama3-8B-Instruct \cite{llama3modelcard} as the vanilla agent and GSM8K as the agent task.
As shown in Figure \ref{fig:specificity}, only 42.61\% tasks could be solved by all fixed single mechanism baselines. This result suggests that more than 50\% of tasks are of mechanism sensitivity. For instance, certain tasks require external knowledge, while others may encounter conflicts upon the introduction of such knowledge. Consequently, we believe that different tasks possess distinct underlying solution structures. Moreover, the oracle mechanism activation results demonstrate that the model can solve 96.89\% of the tasks with the aid of selecting the correct mechanism, highlighting that adaptive mechanism activation has a very high ceiling performance. This suggests a significant potential for identifying the inherent characteristics of tasks and their solution structures. Our ALAMA still falls short of the performance ceiling, which anticipates further optimization of the mechanism activation methods.


\subsection{The Effects of Mixing Different Mechanism Data }
To investigate the impact of individual and mixed mechanisms data on the performance of the agent, we divided $\mathcal{U}_{\textrm{IMAO}}$ and $\mathcal{U}_{\textrm{MAAO}}$ based on different mechanisms. For $\mathcal{U}_{\textrm{MAAO}}$, we segment it according to the mechanisms activated by the positive examples, and incorporated all negative examples of the corresponding tasks into the training set. 
For IMAO, we employed Meta-Llama-3-8B-Instruct as the base model, whereas for MAAO, we utilized ALAMA$_\textrm{IMAO}$ as the base model.

In IMAO, we observed that fine-tuning the model using single-mechanism trajectories leads to underperformance, as the use of original data does not effectively enhance the performance under the zero-shot setting. We hypothesize this may be due to insufficient training data resulting from data segmentation. After sampling more data corresponding to the specific mechanisms for further fine-tuning, it still could not significantly improve the performance of the agent. These performances are shown as 'original' and 'aug' in Table \ref{tab:data}. This suggests that under the single-mechanism activation setting, the quality of trajectories generated through self-exploration is insufficient for the agent to achieve performance comparable to In-Context Learning, and it might require using expert-generated models to attain higher performance. Furthermore, we found that the performance using $\mathcal{U}_{\textrm{IMAO}}$ for training far exceeds that achieved with single-mechanism data, proving the superiority of mixed-mechanism data fine-tuning. In MAAO, the performance using multiple mechanisms for fine-tuning also surpasses that using single-mechanism data. This indicates that an agent has mechanism preferences for different tasks, which aligns with the Residual performance presented in Figure \ref{fig:specificity}. However, the performance gap between full data and partial data is not as pronounced in IMAO as it is in MAAO, suggesting that IMAO plays a more crucial role in agent meta-ability acquisition.

\label{sec:4.2}
\begin{table}[tb]
    \centering
    \resizebox{1\linewidth}{!}{
    \begin{tabular}{lcc}
    \toprule
    Data & Number & Acc\\
    \midrule
    IMAO \\
    \cmidrule{1-1}
    Reason original / aug & 251 / 1300 & 25.47 / 36.01 \\
    Plan original / aug & 264 / 1300 & 28.73 / 36.69 \\
    Memory original / aug & 240 / 1300 & 37.23 / 43.29 \\
    Reflection original / aug & 248 / 1300 & 47.08 / 46.63 \\
    External-Aug original / aug & 254 / 1300 & 37.76 / 43.97 \\
    \midrule
    Full & 1257 & \textbf{78.77} \\
    \midrule
    MAAO \\
    \cmidrule{1-1}
    Reason original & 2403 & 81.43 \\
    Plan original& 2396 & 79.00 \\
    Memory original& 2390 & 78.77 \\
    Reflection original& 2524 & 80.21 \\
    External-Aug original& 1618 & 70.51 \\
    \midrule
    Full & 7120 & \textbf{82.18} \\
    \bottomrule
    \end{tabular}
    }
    \caption{The performance of training agent using different parts of data. Number means the number of the data used in training.}
    \label{tab:data}
\end{table}
\section{Related Work}

\subsection{Language Agent}

To achieve better autonomous task accomplishment, the research community has designed many Language Agent Frameworks like ReAct \cite{yao2023react}, Reflexion \cite{shinn2023reflexion}, and Multi-Agent Debate \cite{du2023improving, liang2023encouraging}. However, these frameworks are labor-intensive for prompt design and work only for big foundation models which are opaque, proprietary, and API-based \cite{chatgpt2022, claude2023}, hindering the research of inherent mechanisms. Another effective technique is adapting open-sourced LLM to LA by imitation fine-tuning \cite{ho-etal-2023-large, zeng2023agenttuning, chen2023fireact, xu2024lemur, yin2024agent, wang2024learning, chen2024agentflan, yin2024lumos}. High-reward trajectories are collected by reformatting golden rationales \cite{samoyed2024} or distilling from ChatGPT \cite{chatgpt2022, chen2023fireact}. These endow smaller models with abilities like planning, reasoning, and reflection. But these LAs are limited as they do not explore the task environments for interactive self-improvement. Exploration fine-tuning \cite{song2024trial, yang2024react, wang2024sotopiapi} has gained attention recently as it shows potential for self-improvement.

\subsection{Self-evolution of Large Language Model}

Self-evolution is crucial for Large Language Models \cite{huang2023large, tao2024survey, lu2024self}. Techniques like ReST (\cite{gulcehre2023reinforced}), self-rewarding (\cite{yuan2024selfrewarding}), and self-play (\cite{chen2024selfplay}) achieve it via iterative generation and optimization. As LLMs evolve beyond human intelligence, more weakly supervised automatic feedback signals are needed for self-evolution (e.g., \cite{burns2023weaktostrong, cao2024scalable}). The approach in this paper is also a method for LLM self-evolution.


\section{Conclusion}
In this paper, we propose \textbf{A}daptive \textbf{L}anguage \textbf{A}gent \textbf{M}echanism \textbf{A}ctivation Learning with Self-Exploration (\textbf{ALAMA}). We observed that numerous tasks exhibit mechanism sensitivity. And the oracle mechanism activation exhibits stronger performance than fixed baselines. To this end, we \textbf{uni}fy different agent mechanisms by \textbf{act}ions (\textbf{UniAct}) into a harmonized agent framework. Moreover, we utilize an adaptive mechanism activation optimization method based on self-exploration, which requires less data than previous SoTA agents and is training-efficient.
Extensive experiments demonstrate the effectiveness and generalization of our proposed method. 
Further analysis shows that increasing the number of mechanisms and integrating trajectory data from different mechanisms are crucial for enhancing agent performance. Code will be available at \texttt{https://github.com/hzy312/alama}.
\section*{Limitations}

In this paper, the discussion of adaptive mechanism activation is limited to the activation of a single mechanism and does not address the simultaneous activation of multiple mechanisms. Activating various mechanisms concurrently could offer additional benefits; however, it also increases the complexity of learning adaptive mechanism activation. Therefore, we consider this an area for future work to be explored subsequently. Moreover, in Section \ref{sec:4.2}, we discuss only the effects of full data and single-mechanism data, omitting the impact of mixing data from different mechanisms. The five mechanisms discussed in this paper could lead to $2^5 - 1$ possible combinations, and our limited computational resources did not allow for the evaluation of all possibilities. We plan to incorporate these data in a formal version later for further discussion.

\bibliography{custom}

\appendix
\newpage

\section{Data}
\label{app:data}
\begin{table}[htbp]
    \centering
    \begin{tabular}{lcc}
    \toprule
      Dataset  & \#Train & \#Test \\
      \midrule
      GSM8K  & 7473 & 1319 \\
      NumGLUE & 0 & 254 \\
      SVAMP & 0 & 1000 \\
      HotpotQA & 10000 & 500 \\
      TriviaQA & 0  &  500\\
      Bamboogle & 0 & 125 \\
      \bottomrule
    \end{tabular}
    \caption{The statistic of data used in our experiments.}
    \label{tab:my_label}
\end{table}
 For datasets with large test sets, we perform down-sampling. Furthermore, to increase the difficulty of the test sets, we filter out some relatively simpler data points in some datasets. For HotpotQA, we have filtered out questions that can be answered with "yes" or "no", and then sample 10000 from the train split. For HotpotQA and TriviaQA, we have sampled 500 questions from the dev split as the test set.

\section{Training and Inference}
\label{app:train}

\begin{table}[htbp]
    \centering
    \begin{tabular}{lc}
    \toprule
    IMAO \\
    \cmidrule{1-1}
    Key     & Value  \\
    epoch   & 4 \\
    batch size & 8\\
    learning rate & 1e-6 \\
    learning rate scheduler & cosine \\
    warmup ratio & 0.1 \\
    \midrule 
    MAAO \\
    \cmidrule{1-1}
    Key     & Value  \\
    epoch   & 2 \\
    batch size & 16\\
    learning rate & 1e-7 \\
    learning rate scheduler & cosine \\
    warmup ratio & 0.1 \\
    $\frac{\lambda_D n_D}{\lambda_U n_U}$ & 4/3 \\
    \bottomrule
    \end{tabular}
    \caption{Hyperparameters for training.}
    \label{tab:hyp}
\end{table}

 For LLMs training, we employ TRL \cite{vonwerra2022trl} and Deepspeed \cite{10.1145/3394486.3406703} as the frameworks to conduct full fine-tuning. Due to the limited availability of our computational resources, we utilize Zero3+offload \cite{273920} during the fine-tuning process. The hyperparameters are listed in \ref{tab:hyp}. For LLMs inference, we utilize vllm \cite{10.1145/3600006.3613165} for acceleration.

\section{Algorithm}
\begin{algorithm}[htb]
\caption{ALAMA: Adaptive Language Agent Mechanism Activation with Self-Exploration}
\label{algo:alama}
\begin{algorithmic}[1]

\Require $\mathcal{M} = \{m_i\}_{i=1}^{5}$; $\mathcal{D} = \{d_i\}_{i=1}^{5}$; $\mathcal{T} = \{t_j\}_{j=1}^{|\mathcal{T}|}$; LA$_{\theta}$ 
\State $\mathcal{U, R} \gets \emptyset$
\Comment{Initialize UniAct Trajectory and Reward set}
\For{$i \gets 1$ to 5}
\Comment{Self-Exploration}
\For{$j \gets 1$ to $\mathcal{T}$}
  \State $s_{i,j}, r_{i,j} \gets \textrm{LA}_{\theta}(d_i, t_j)$ 
  \State   $u_{i, j}$  $\gets$  $\textrm{UniActTrans}(s_{i,j})$
  \State  $\mathcal{U}$.append($u_{i,j}$), $\mathcal{R}$.append($r_{i,j}$)
\EndFor
\EndFor
\State $\mathcal{U}_{\textrm{IMAO}}, \mathcal{U}_{\textrm{MAAO-pos}}, \mathcal{U}_{\textrm{MAAO-neg}} \gets \emptyset$
\item[] \Comment{Initialize IMAO set and MAAO set}
\For{$j \gets 1$ to $\mathcal{T}$}
\If{$\forall i \in [1,5], r_{i,j} = 1$}
\State pass
\Else
\For{$i \gets 1$ to $5$}
\If{$r_{i,j} == 1$ }
\State $\mathcal{U}_{\textrm{MAAO-pos}}$.append($u_{i,j}$)
\Else
\State $\mathcal{U}_{\textrm{MAAO-neg}}$.append($u_{i,j}$)
\EndIf
\EndFor
\EndIf
\EndFor

\State $\mathcal{U}_{\textrm{IMAO}} \gets$ $\mathcal{U}_{\textrm{MAAO-pos}}$
\State Update LA$_{\theta}$ with Implicit Mechanism Activation Optimization $\mathcal{L}_{\textrm{IMAO}}$ on $\mathcal{U}_{\textrm{IMAO}}$
\State Update LA$_{\theta}$ with Mechanism Activation Adaptability Optimization $\mathcal{L}_{\textrm{MAAO}}$ on $\mathcal{U}_{\textrm{MAAO}}$
\State \Return $\textrm{LA}_{\textrm{final}}$

\end{algorithmic}
\end{algorithm}

\section{Implementation of Different Mechanisms}
\label{implementation}
Existing works have significantly enhanced the ability of LLM to solve different tasks through different prompting methods. For example, CoT \cite{wei2022chain} can improve reasoning ability, and Reflexion \cite{shinn2023reflexion} can enhance the ability to find errors and self-repair. These different prompting methods can endow the Agent based on LLM with different capabilities to adapt to different task environments. We regard these different capabilities as different mechanisms of the Agent and believe that endowing the Language Agent with different mechanisms can bring different benefits for performance improvement. We use In-Context Learning to activate the corresponding mechanism. Below, we will map the mechanisms to the corresponding prompting methods to show how to implement them and clarify the benefits brought by different mechanisms.

\texttt{Reason -> CoT} \cite{wei2022chain}: Chain-of-thought significantly improves the performance of the model in downstream tasks by explicitly making the model generate the reasoning process. This prompting method can endow the Language Agent with the reasoning ability.

\texttt{Plan -> Plan-and-Solve} \cite{wang-etal-2023-plan}: Plan-and-Solve first decomposes the task and then solves the sub-tasks step by step to obtain the final answer. This method can decompose difficult tasks into multiple simple and easy-to-solve tasks to improve performance. This prompting method can enhance the planning and task decomposition ability of the Language Agent.

\texttt{Memory -> ExpNote} \cite{sun2023expnote}: We first inference on the training set of the Held-in tasks with CoT method and collect all the wrong trajectories, treating all these errors as a wrong-answer notebook. During testing, we search in the wrong-answer notebook, retrieve similar problems, and explicitly prompt the LLM not to make similar mistakes. We use the text-embedding-3-small\footnote{https://platform.openai.com/docs/guides/embeddings/embedding-models} from OpenAI as the embedding model. This prompting method can enhance the ability of the Language Agent to utilize past experience.

\texttt{Reflection -> Reflexion} \cite{shinn2023reflexion}: Reflexion finds and corrects possible errors in the previous steps through the reflection method. It is well belived that self-generation reflection \cite{huang2024large} might deteriorate the performance, so we choose the Deepseek-V2 \cite{deepseekai2024deepseekv2} as the expert Critic Model. This prompting method can enhance the ability of the Language Agent to find errors and self-repair.

\texttt{External-Augmentation -> ReAct} \cite{yao2023react}: This method gives LLM the ability to call tools and borrow external capabilities to improve the performance of the model. For example, a calculator can be called in math tasks, and a search engine can be called in knowledge-intensive reasoning tasks. This prompting method can significantly expand the ability boundary of the Language Agent.

\section{UniActTransform}
\label{app:transform}
The corresponding extracted contents descripted below are filled into the UniAct format in Appendix \ref{sec:prompt}.

\texttt{Reason}: We extract the thought and answer from the ICL trajectories and fill them into the UniAct format.

\texttt{Plan}: We extract the plan, thought and answer from the ICL trajectories and fill them into the UniAct format.

\texttt{Memory}: We retrieve the failed case and extract the thought and answer from the ICL trajectories and fill them into the UniAct format.

\texttt{Reflection} We extract the first-generated thought, reflection reviews from the expert Critic model, and second-generated thought and corresponding answer to fill into the UniAct format.

\texttt{External Augmentation}: We extract the external tool output (calculator results or search engine results) to fill into the UniAct format.

\section{Prompt of UniAct}
\label{sec:prompt}

We show the UniAct format template used in this paper. We show the system, \texttt{Reason}, \texttt{Plan}, \texttt{Memory}, \texttt{Reflection}, \texttt{External-Augmentation} prompt for mathmetical reansoning and knowledge-intensive reasoning tasks in Table 6-11 and Table 12-17.

\begin{table*}[htbp]
\centering
\begin{tabularx}{\textwidth}{X}
\toprule 
\textbf{system} \\
\cmidrule{1-1}
You are an agent that has five important mechanisms for solving a problem: Reason, Plan, Augmentation, Reflection, Memory.\\

Reason: The agent will do reasoning to solve a problem step by step.\\
Plan: The agent will devise a detailed plan and then carry out the plan step by step to solve the problem\\
Augmentation: The agent will interleave the reasoning and action to solve the problem. The action will call the Calculator for more precise numerical calculation.\\
Reflection: After reasoning, the agent will reflect on the previous reasoning and corresponding answer and get critic reviews. Based on the reviews, the agent will refine its reasoning and answer again.\\
Memory: The agent has a memory database of failed reasoning trajectories. For each question, the agent will retrieve failed case from the memory as the reference to avoid such type of errors.\\

You can use these mechanisms to solve problems.\\
You have to think and solve the problem step-by-step with interleaving Thought, Action, Observation steps.\\
Thought is your reasoning process.\\
Action could be:\\
    $-$ Make plan: The agent will devise a detailed plan and then carry out the plan step by step to solve the problem.\\
    $-$ Carry out plan: The agent will carry out the plan step by step to solve the problem.\\
    $-$ Reflect: The agent will reflect on the previous reasoning and corresponding answer and get critic reviews. Based on the reviews, the agent will refine its reasoning and answer again.\\
    $-$ Retrieve memory: The agent will retrieve failed case from the memory as the reference to avoid such type of errors.\\
    $-$ Calculate: The agent will call the Calculator for more precise numerical calculation. Using format: Calculate[the expression]\\
    $-$ Finish: The agent will give the final answer.\\
Observation is the feedback from the environment.\\

The answer needs to be a number and cannot contain other characters.\\

Your response should use the following format:\\

Thought: <your thoughts>\\
Action: <your next action>  \\
\bottomrule
\end{tabularx}
\caption{System prompt of UniAct for Mathematical Reasoning.}
\label{tab:my_label}
\end{table*}

\begin{table*}[htbp]
\centering
\begin{tabularx}{\textwidth}{X}
\toprule 
\cmidrule{1-1}
\textbf{Reason} \\ 
\cmidrule{1-1}
\texttt{Environment}: Task: \{task\} \\
\texttt{Agent}: Thought: \{thought\} Action: Finish[\{answer\}] \\
\bottomrule
\end{tabularx}
\caption{\texttt{Reason} format of UniAct for Mathematical Reasoning.}
\label{tab:my_label}
\end{table*}

\begin{table*}[htbp]
\centering
\begin{tabularx}{\textwidth}{X}
\toprule 
\cmidrule{1-1}
\textbf{Plan} \\
\cmidrule{1-1}
\texttt{Environment}: Task: \{task\} \\
\texttt{Agent}: Thought: Devising a detailed plan before solving this problem may be helpful. Action: Make plan \\
\texttt{Environment}: Observation: Let's first understand the problem, extract relevant variables and their corresponding numerals, and make a complete plan. \\
\texttt{Agent}: Thought: My plan: \{plan\} Action: Carry out plan \\
\texttt{Environment}: Observation: Let's carry out the plan, calculate intermediate variables (pay attention to correct numerical calculation and commonsense), solve the problem step by step. \\
\texttt{Agent}: Thought: \{thought\} Action: Finish[\{answer\}]  \\
\bottomrule
\end{tabularx}
\caption{\texttt{Plan} format of UniAct for Mathmetical Reasoning.}
\label{tab:my_label}
\end{table*}

\begin{table*}[htbp]
\centering
\begin{tabularx}{\textwidth}{X}
\toprule 
\cmidrule{1-1}
\textbf{Memory} \\
\cmidrule{1-1}
\texttt{Environment}: Task: \{task\} \\
\texttt{Agent}: Thought: Based on the give question, I will retrieve from my memory to get a similar question and the corresponding wrong solution as the reference to avoid such type of errors and solve the given problem step by step. Action: Retrieve memory \\ 
\texttt{Environment}: Observation: Case: \{case\}. Try to avoid such types of errors. \\
\texttt{Agent}: Thought: \{thought\} Action: Finish[\{answer\}] \\
\bottomrule
\end{tabularx}
\caption{\texttt{Memory} format of UniAct for Mathematical Reasoning.}
\label{tab:my_label}
\end{table*}

\begin{table*}[htbp]
\centering
\begin{tabularx}{\textwidth}{X}
\toprule 
\cmidrule{1-1}
\textbf{Reflection} \\
\cmidrule{1-1}
\texttt{Environment}: Task: \{task\} \\
\texttt{Agent}: Thought: \{pre thought\} Action: Reflect \\
\texttt{Environment}: Observation: Reflection: \{reflection\} Based on the reflection reviews, please refine the thought and action. \\
\texttt{Agent}: Thought: \{post thought\} Action: Finish[\{answer\}] \\
\bottomrule
\end{tabularx}
\caption{\texttt{Reflection} format of UniAct for Mathematical Reasoning.}
\label{tab:my_label}
\end{table*}

\begin{table*}[htbp]
\centering
\begin{tabularx}{\textwidth}{X}
\toprule 
\cmidrule{1-1}
\textbf{External Augmentation} \\
\cmidrule{1-1}
\texttt{Environment}: Task: \{task\} \\
\texttt{Agent}: Thought: \{thought\} Action: Calculate[\{expression\}]\\
\texttt{Environment}: Observation: \{result\}\\
... \\
\texttt{Agent}: Thought: \{thought\} Action: Finish[\{answer\}] \\
\bottomrule
\end{tabularx}
\caption{\texttt{External Augmentation} format of UniAct for Mathematical Reasoning.}
\label{tab:my_label}
\end{table*}

\begin{table*}[htbp]
\centering
\begin{tabularx}{\textwidth}{X}
\toprule 
\textbf{system} \\
\cmidrule{1-1}
You are an agent that has five important mechanisms for solving a problem: Reason, Plan, Augmentation, Reflection, Memory.\\

Reason: The agent will do reasoning to solve a problem step by step.\\
Plan: The agent will devise a detailed plan and then carry out the plan step by step to solve the problem\\
Augmentation: The agent will interleave the reasoning and action to solve the problem. The action will call the Wikipedia Search for more precise knowledge. \\
Reflection: After reasoning, the agent will reflect on the previous reasoning and corresponding answer and get critic reviews. Based on the reviews, the agent will refine its reasoning and answer again.\\
Memory: The agent has a memory database of failed reasoning trajectories. For each question, the agent will retrieve failed case from the memory as the reference to avoid such type of errors.\\

You can use these mechanisms to solve problems.\\
You have to think and solve the problem step-by-step with interleaving Thought, Action, Observation steps.\\
Thought is your reasoning process.\\
Action could be:\\
    $-$ Make plan: The agent will devise a detailed plan and then carry out the plan step by step to solve the problem.\\
    $-$ Carry out plan: The agent will carry out the plan step by step to solve the problem.\\
    $-$ Reflect: The agent will reflect on the previous reasoning and corresponding answer and get critic reviews. Based on the reviews, the agent will refine its reasoning and answer again.\\
    $-$ Retrieve memory: The agent will retrieve failed case from the memory as the reference to avoid such type of errors.\\
    $-$ Search, which searches the exact entity on Wikipedia and returns the first paragraph if it exists. If not, it will return some similar entities to search. Using format: Search[entity] \\
    $-$ Lookup, which returns the next sentence containing keyword in the current passage. Using format: Lookup[keyword] \\
    $-$ Finish: The agent will give the final answer.\\
Observation is the feedback from the environment.\\

Your response should use the following format:\\

Thought: <your thoughts>\\
Action: <your next action>  \\
\bottomrule
\end{tabularx}
\caption{System prompt of UniAct for Knowledge-intensive Reasoning.}
\label{tab:my_label}
\end{table*}

\begin{table*}[htbp]
\centering
\begin{tabularx}{\textwidth}{X}
\toprule 
\cmidrule{1-1}
\textbf{Reason} \\ 
\cmidrule{1-1}
\texttt{Environment}: Task: \{task\} \\
\texttt{Agent}: Thought: \{thought\} Action: Finish[\{answer\}] \\
\bottomrule
\end{tabularx}
\caption{\texttt{Reason} format of UniAct for Knowledge-intensive Reasoning.}
\label{tab:my_label}
\end{table*}

\begin{table*}[htbp]
\centering
\begin{tabularx}{\textwidth}{X}
\toprule 
\cmidrule{1-1}
\textbf{Plan} \\
\cmidrule{1-1}
\texttt{Environment}: Task: \{task\} \\
\texttt{Agent}: Thought: Devising a detailed plan before solving this problem may be helpful. Action: Make plan \\
\texttt{Environment}: Observation: Let's first understand the problem, decompose the question if necessary, and make a complete plan. \\
\texttt{Agent}: Thought: My plan: \{plan\} Action: Carry out plan \\
\texttt{Environment}: Observation: Let's carry out the plan, get the intermediate answers explicitly step-by-step, and integrate these evidences to get the final anwer. \\
\texttt{Agent}: Thought: \{thought\} Action: Finish[\{answer\}]  \\
\bottomrule
\end{tabularx}
\caption{\texttt{Plan} format of UniAct for Knowledge-intensive Reasoning.}
\label{tab:my_label}
\end{table*}

\begin{table*}[htbp]
\centering
\begin{tabularx}{\textwidth}{X}
\toprule 
\cmidrule{1-1}
\textbf{Memory} \\
\cmidrule{1-1}
\texttt{Environment}: Task: \{task\} \\
\texttt{Agent}: Thought: Based on the given question, I will retrieve from my memory to get a similar question and the corresponding wrong solution as the reference to avoid such types of errors and solve the given problem step by step. Action: Retrieve memory \\ 
\texttt{Environment}: Observation: Case: \{case\}. Try to avoid such types of errors. \\
\texttt{Agent}: Thought: \{thought\} Action: Finish[\{answer\}] \\
\bottomrule
\end{tabularx}
\caption{\texttt{Memory} format of UniAct for Knowledge-intensive Reasoning.}
\label{tab:my_label}
\end{table*}

\begin{table*}[htbp]
\centering
\begin{tabularx}{\textwidth}{X}
\toprule 
\cmidrule{1-1}
\textbf{Reflection} \\
\cmidrule{1-1}
\texttt{Environment}: Task: \{task\} \\
\texttt{Agent}: Thought: \{pre thought\} Action: Reflect \\
\texttt{Environment}: Observation: Reflection: \{reflection\} Based on the reflection reviews, please refine the thought and action. \\
\texttt{Agent}: Thought: \{post thought\} Action: Finish[\{answer\}] \\
\bottomrule
\end{tabularx}
\caption{\texttt{Reflection} format of UniAct for Knowledge-intensive Reasoning.}
\label{tab:my_label}
\end{table*}

\begin{table*}[htbp]
\centering
\begin{tabularx}{\textwidth}{X}
\toprule 
\cmidrule{1-1}
\textbf{External Augmentation} \\
\cmidrule{1-1}
\texttt{Environment}: Task: \{task\} \\
\texttt{Agent}: Thought: \{thought\} Action: Search[\{entity\}] or Lookup[\{keyword\}]\\
\texttt{Environment}: Observation: \{result\}\\
... \\
\texttt{Agent}: Thought: \{thought\} Action: Finish[\{answer\}] \\
\bottomrule
\end{tabularx}
\caption{\texttt{External Augmentation} format of UniAct for Knowledge-intensive Reasoning.}
\label{tab:my_label}
\end{table*}

\end{document}